\documentclass[aps, reprint]{revtex4-2}
\usepackage{graphicx}
\usepackage{bm}
\usepackage{amsmath}
\usepackage{amssymb}
\usepackage{physics}
\usepackage{xcolor}
\usepackage[capitalize]{cleveref}
\usepackage{algorithm2e}
\usepackage{algpseudocode}
\newtheorem{definition}{Definition}

\begin{document}

\title{Leveraging Trust for Joint Multi-Objective and Multi-Fidelity Optimization}
\author{Faran Irshad}
\author{Stefan Karsch}
\author{Andreas D\"opp}
\affiliation{Faculty of Physics, Ludwig Maximilian University of Munich, Am Coulombwall 1, 85748 Garching, Germany}

\begin{abstract}
In the pursuit of efficient optimization of expensive-to-evaluate systems, this paper investigates a novel approach to Bayesian multi-objective and multi-fidelity (MOMF) optimization. Traditional optimization methods, while effective, often encounter prohibitively high costs in multi-dimensional optimizations of one or more objectives. Multi-fidelity approaches offer potential remedies by utilizing multiple, less costly information sources, such as low-resolution simulations. However, integrating these two strategies presents a significant challenge. We suggest the innovative use of a trust metric to support simultaneous optimization of multiple objectives and data sources. Our method modifies a multi-objective optimization policy to incorporate the trust gain per evaluation cost as one objective in a Pareto optimization problem, enabling simultaneous MOMF at lower costs. We present and compare two MOMF optimization methods: a holistic approach selecting both the input parameters and the trust parameter jointly, and a sequential approach for benchmarking. Through benchmarks on synthetic test functions, our approach is shown to yield significant cost reductions - up to an order of magnitude compared to pure multi-objective optimization. Furthermore, we find that joint optimization of the trust and objective domains outperforms addressing them in sequential manner. We validate our results for the use case of optimizing laser-plasma acceleration simulations, demonstrating our method's potential in Pareto optimization of high-cost black-box functions. Implementing these methods in existing Bayesian frameworks is simple, and they can be readily extended to batch optimization. With their capability to handle various continuous or discrete fidelity dimensions, our techniques offer broad applicability in solving simulation problems in fields such as plasma physics and fluid dynamics.
\end{abstract}

\maketitle

\section{Introduction}
\label{intro}

Optimizing black-box functions, which involve systems where only the input-output relationship is considered without knowledge of the internal workings, is crucial in various fields such as computer science, engineering, and physics \cite{snoek2012practical,mockus1994application,jones1998efficient,packwood2017bayesian,shoemaker2007watershed, dopp_2023}. While these problems are well understood for some kinds of functions, there exist various properties that can make black-box functions very difficult to optimize. One of the challenges popularly known as the ``curse of dimensionality,'' that makes optimization increasingly difficult as the number of input dimensions grows. 
Moreover, many optimization problems involve multiple objectives, which can be challenging to define or even conflicting in nature. These kind of multi-objective optimization problems arise in diverse research domains \cite{sharma2022comprehensive} including but not restricted to design of photonic crystal filters \cite{mirjalili2017multi}, aerospace design problems \cite{arias2012multiobjective} and robotics \cite{avder2019multi}. Multi-objective problems can be solved by directly examining different trade-offs between various objectives and selecting an optimal combination that leads to the desired outcome. This concept is embodied by Pareto efficiency, visualized as the Pareto front, which represents a set of optimal trade-offs between competing objectives. Finding this Pareto front is usually referred to as Multi-objective optimization (MOO) and various techniques have been developed to tackle them \cite{branke2008}. Among the most popular are techniques based on evolutionary algorithms and, more recently, Bayesian optimization. While evolutionary algorithms focus on the efficient combinations of successful ("fit") parameters, Bayesian techniques try to approximate the unknown black box function with fast surrogate models for its mean and variance. Incorporating multiple information sources into the optimization process can further improve efficiency, especially when dealing with costly black-box functions. Such information sources or 'fidelities' could for instance be simulations with different numerical resolution leading to approximations of the underlying physical processes. Multi-fidelity optimization can leverage information from different sources, each with varying levels of fidelity, accuracy, or computational cost \cite{klein2017fast}. Using this information, multi-fidelity optimization can provide more informed decision-making while reducing the overall computational burden.

In this article, we detail an approach that expands the scope of multi-objective optimization to encompass varied fidelity levels, allowing us to build the Pareto front more economically using cost-efficient approximations of the core black box function. Central to our method is the inclusion of an objective that embodies the information content—termed "trust"—which can be seamlessly optimized alongside other objectives using proven numerical strategies from the realm of multi-objective optimization. The manuscript is structured as follows: After an introduction to Bayesian optimization via Gaussian process regression (2.1), we discuss common acquisition functions used for single-objective, single-fidelity optimization (2.2). We then outline how these policies are generalized to either multi-fidelity (2.3) and multi-objective (2.4) optimization. In Section 3 we then use these concepts to devise policies to perform combined multi-fidelity, multi-objective optimization, either in a joint fashion via trust (3.1) or in a sequential manner (3.2). We benchmark our results for different test functions to understand the behaviour of proposed multi-objective multi-fidelity algorithms (3.3). We then apply the single step algorithm to a real world application. The example is taken from computational physics where simulations are optimized (3.4). Both test functions and the computational physics application illustrate significantly faster convergence times towards the Pareto front than for Bayesian optimization without multiple fidelities. In Section 4 we summarize our results and give an outlook on potential applications.

\begin{figure*}[t]
    \centering
    \includegraphics[trim=60 0 60 0,clip,width=1\textwidth]{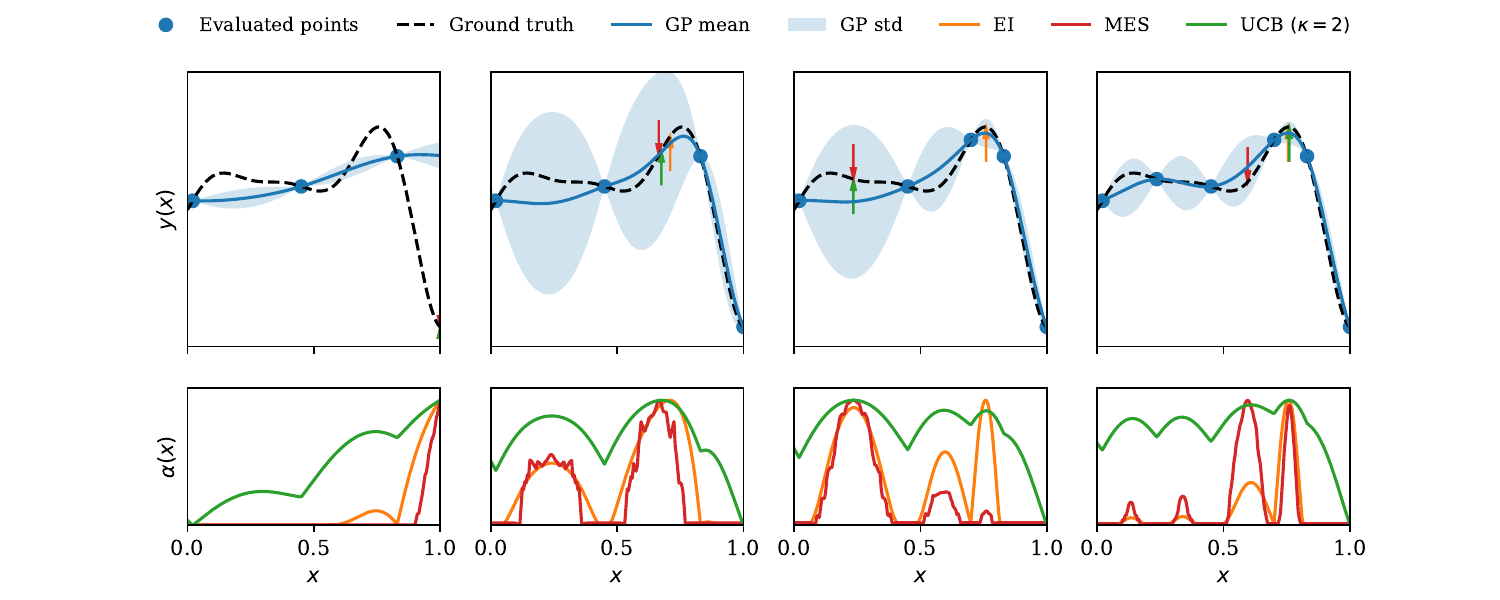}
    \caption{Consecutive iterations of Bayesian optimization (maximization) of the Forrester test function. \textit{Top}: The dashed black line represents the true function that is being optimized. The blue line is the GP regression mean with a shaded region around it representing the standard deviation. The blue dots are the points that were evaluated. Arrows indicate the next measurement point suggested by the respective acquisition functions from the bottom plot, showing the different prioritization towards exploration and exploitation. \textit{Bottom}: Evaluations of three different metrics, expected improvement (EI, Eq. \ref{EI}), max-value entropy search (MES, Eq.\ref{MES}) and upper confidence bound (UCB, Eq. \ref{UCB} with $\kappa=2$) for each step in the upper plot.}
    \label{BO_visualization}
\end{figure*}

\section{Background and Related Work}
\label{sec:2}

In this section we provide an overview of the key components of Bayesian optimization, including Gaussian processes and acquisition functions, and discuss their extension to multi-objective or multi-fidelity optimization.
The models underlying most Bayesian optimization are Gaussian processes (GPs), which describe a black-box objective function as a distribution over functions in a continuous domain \cite{rasmussen2003gaussian}. GPs consist of a mean function, which represents the average value of the objective function, and a kernel function, which captures the relationship between different points in the parameter space. Prior information about the function can be incorporated either in the mean function or the kernel. In many cases the prior mean is assumed to be zero throughout the parameter space and all the prior information is encoded in the kernel function. As the black-box function is evaluated at specific positions, the GP is updated based on these evaluations. In essence, the GP combines the prior information about the objective function with the evaluated data resulting in a so called Posterior distribution. A function usually denoted as an acquisition function uses this posterior distribution to predict next possible best position to evaluate the objective function. By optimizing the acquisition function, which is computationally efficient, the problem of optimizing the expensive black-box function is transformed into optimizing the acquisition function.

\subsection{Acquisition Functions for Single-Objective, Single-Fidelity Optimization}
\label{sec:2.2}

Acquisition functions encompass a broad class of functions that can be constructed using the posterior distribution of the GP model. The effectiveness of an acquisition function is assessed based on its ability to converge to the global optimum, with minimal evaluations of the objective function. This can be best illustrated through a series of diagrams shown in Figure \ref{BO_visualization}, depicting the iterations of Bayesian optimization loop starting from three initial points. After constructing a GP model with a mean (solid blue line) and variance (shaded region), an acquisition function is derived (bottom plot). This acquisition function is then optimized, often using gradient methods, to identify its maximum, which corresponds to the next evaluation point for the black-box objective function. The choice of the acquisition function significantly impacts the convergence of Bayesian optimization towards the global optimum.

This section provides an overview of widely recognized acquisition function policies discussed in the existing literature. Assuming that the objective function has been evaluated $n$ times, these acquisition functions propose the optimal choice for the subsequent evaluation point, denoted as $x_{n+1}$. This selection is achieved through the optimization of a derived metric, taking into account the information from previously evaluated points. Specifically, the next evaluation point is given by $x_{n+1} = \arg\max_{x\in\mathcal{X}}D^n(x)$, where $\mathcal{X}$ is the input parameter domain and $D^n$ represents one of the acquisition functions detailed below.

One widely used acquisition function is the upper confidence bound (UCB) policy \cite{cox1992}, which is alternatively referred to as the lower confidence bound for minimization tasks. The UCB acquisition function is expressed as:

\begin{equation}
    \text{UCB}(x) = \mu(x) + \kappa\sigma(x),
    \label{UCB}
\end{equation}

where $\mu(x)$ and $\sigma(x)$ denote the mean and standard deviation, respectively, derived from Gaussian processes. The hyperparameter $\kappa$ is used to balance exploration and exploitation. UCB is popular due to its simplicity and effectiveness in practice.

Another well-known acquisition function in Bayesian optimization is Expected Improvement (EI) \cite{mockus1994}, which suggests the next evaluation point based on the expected improvement over the current optimal objective value $y^*$. The EI acquisition function is expressed as:

\begin{equation}
    \text{EI}_n = \mathbb{E}[\max(f_{n+1}(x) - y^*, 0)],
    \label{EI}
\end{equation}

where $f_{n+1}(x)$ is calculated from the posterior distribution. A variation of EI is the Knowledge Gradient (KG) acquisition function \cite{frazier2008, scott2011}, which relies entirely on the posterior model. KG selects the next evaluation point based not on the best observable value but on the best value of the posterior mean. The KG acquisition function is given by:

\begin{equation}
    \text{KG}_n = \mathbb{E}[\max_{x\in\mathcal{X}}(\mu_{n+1}(x)) - \max_{x\in\mathcal{X}}(\mu_{n}(x))],
    \label{KG}
\end{equation}

where $\mu(x)$ is the posterior mean of the GP. KG is more exploratory than EI as it is influenced by posterior changes throughout the domain.

Information-theoretic acquisition functions utilize the mutual information $I(x, x^*)$ between a specific location in the parameter domain and the observed data set. One such function is Entropy Search (ES) \cite{hennig2012}, represented mathematically as:

\begin{align}
    \text{ES}_n &= I([x_{n+1},y_{n+1}];x^*|[x_n,y_n]) \\&= H(p_n(x^*)) - \mathbb{E}[H(p_n(x^*|[x_{n+1},y_{n+1}]))],
    \label{ES}
\end{align}

where $H$ denotes Shannon's entropy and $p(x)$ refers to the posterior distribution. The left term in the equation represents the entropy of the posterior distribution of the maximizing location $x^*$, while the right term depicts an expectation over the entropy of the posterior after an additional sample. In higher-dimensional input spaces, evaluating the mutual information between the point to be queried and the maximizing location $x^*$ becomes challenging. To address this, computationally more efficient variations have been introduced. Max-value entropy search (MES) \cite{wang2017} utilizes the mutual information between the maximum value $y^*$ rather than $x^*$. The MES acquisition function is formulated as:

% \begin{widetext}
% \begin{equation}
%     \begin{aligned}
%         \text{MES}_n &= I([x_{n+1},y_{n+1}];y^*|[x_n,y_n]) \\
%         &= H(p_n(y^*|[x_{n},y_{n}],x_{n+1})) - \mathbb{E}[H(p_n(y|[x_{n},y_{n}],x_{n+1},y^*))].
%     \end{aligned}
%     \label{MES}
% \end{equation}
% \end{widetext}
\begin{equation}
    \begin{split}
        \text{MES}_n &= I([x_{n+1},y_{n+1}];y^*|[x_n,y_n]) \\
        &= H(p_n(y^*|[x_{n},y_{n}],x_{n+1})) \\&- \mathbb{E}[H(p_n(y|[x_{n},y_{n}],x_{n+1},y^*))].
    \end{split}
    \label{MES}
\end{equation}

MES offers comparable or even better performance than ES while being significantly faster to compute \cite{wang2017}.

\begin{figure*}[t]
  \centering
  \includegraphics*[width=.95\linewidth]{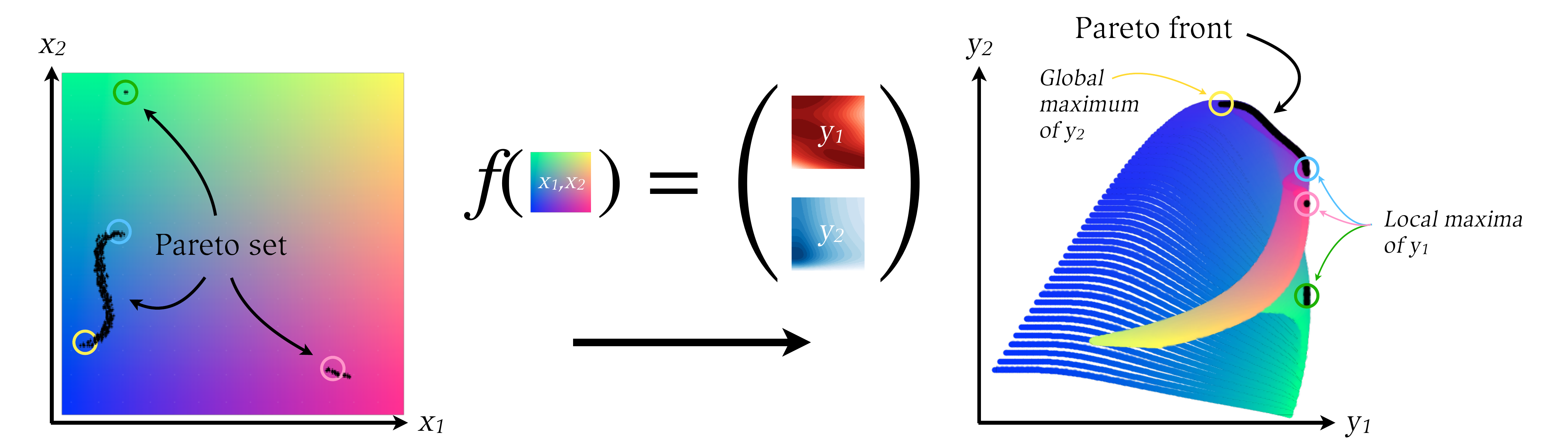}
  \caption{Pareto front. Illustration how a multi-objective function $\bm f(\bm x)=\bm y$ acts on a two-dimensional input space $\bm x = (x_1,x_2)$ and transforms it to the objective space $\bm y = (y_1,y_2)$ on the right. The entirety of possible input positions is uniquely color-coded on the left and the resulting position in the objective space is shown in the same color on the right. The Pareto front is the ensemble of points that dominate others, meaning points that give the highest combination of $y_1$ and $y_2$. The corresponding set of coordinates in the input space is called the Pareto set. Note that both the Pareto front and the Pareto set may be continuously defined locally, but can also contain discontinuities when local maxima get involved. In this example, $f$ is a modified version of the Branin-Currin function from \cite{dixon1978,currin1991} that exhibits a single, global maximum in $y_2$ but multiple local maxima in $y_1$, see also illustration in the center.}
  \label{Pareto}
\end{figure*}

\subsection{Multi-Objective, Single-Fidelity Optimization}
\label{sec:2.4}
As we outlined in the introduction, there exist many use cases in which the objective consists of multiple sub-goals. One can think of these sub-goals as a vector of solutions, which has to be reduced to a scalar number to be compatible with conventional numerical optimization schemes. This reduction, or scalarization, is often not straightforward and the weights given to each sub-goal are usually found empirically. A potent strategy for such use cases is multi-objective optimization, where one tries to increase the diversity of solutions. In this case the function that is being maximized can be expressed as a vector of functions
$$\bm f(\bm x) =\begin{pmatrix}{ f_1({\bm x})}
\\f_2(\bm x) \\ \dots
\end{pmatrix} $$ 
that are evaluated to yield output vectors $$\bm y(\bm x)=\begin{pmatrix}{ y_1({\bm x})}
\\y_2(\bm x) \\ \dots
\end{pmatrix}.$$
The optimum solution in this case consists of a solution vector. A useful concept describing this situation of not just a single optimum but a set of optimal points in the multi-dimensional objective space is the \emph{Pareto efficiency} \cite{branke2008}, which is visualized as the \emph{Pareto front} ($\mathcal{P}$). To describe the Pareto front, the notion of \emph{domination} is outlined in Appendix: Definition \ref{domination} with two objective functions ($f_1(x),f_2(x)$). The Pareto front is composed of a set of \emph{non-dominated} points in the output space as shown in Fig. \ref{Pareto}. 

One of the simplest algorithms for multi-objective optimization is ParEGO (Pareto Efficient Global Optimization) proposed by Knowles \cite{knowles2006parego}. ParEGO is a scalarization-based approach in which a single-objective optimizer is employed to find the Pareto front. In this algorithm, scalarization is achieved using a weighted sum of the objective functions with random weights generated at each iteration. The main advantage of ParEGO over other scalarization methods is that the random weights allow for exploring the Pareto front for a diverse set of solutions without predefining the weights or relying on user preferences. By iteratively running this algorithm with new weights, it becomes possible to approximate the entire Pareto optimal set. 

Alternatively, one can reframe the multi-objective problem as a single-objective optimization by using a new objective that implicitly increases solution diversity. One such criterion is the Hypervolume Indicator. Hypervolume (HV) is defined as the n-dimensional volume of the output subspace covered from a reference point, always taken to be zero in this work, to a set of points in the objective space. Using this definition of HV, we can then proceed to define Hypervolume Improvement (HVI) as
\begin{equation}
    \mbox{HVI}
    (\mathcal{P},y)=\mbox{HV}(\mathcal{P} \cup {y}) - \mbox{HV} (\mathcal{P})  
    \label{HVI}
\end{equation}   
Equation \ref{HVI} describes the difference between the current hypervolume and one with an additional output point $y$ \cite{yang2019multi}. If the set of points making up $\mathcal{P}$ already dominate $y$ then $HVI = 0$, because there is no hypervolume gained by adding the point $y$. 

HVI can be used to generalize the expected improvement policy described in Section 1.2 to the multi-objective scenario. First proposed by Emmerich et al. \cite{emmerich2006single}, this method is called \emph{Expected Hypervolume Improvement (EHVI)}. Following the definition from Yang et al. \cite{yang2019multi}, we can write this as
\begin{equation}
    \mbox{EHVI}(\mu,\sigma,\mathcal{P},y)=E[\mbox{HVI}]= \int \mbox{HVI}(\mathcal{P},y).\mbox{PDF}_{\mu,\sigma}(y) \,dy,  
    \label{EHVI}
\end{equation}
where the probability density function PDF$_{\mu,\sigma}$ is the independent multivariate normal distribution with mean $\mu$ and standard deviation $\sigma$. This infill criterion has been demonstrated to achieve a good convergence to the true Pareto front \cite{couckuyt2014fast,luo2014kriging,shimoyama2013kriging}.

A common criticism of EHVI acquisition function has been the time complexity involved in calculating it. A first closed form calculation of EHVI was implemented by Emmerich et al. \cite{emmerich2011} with a computational complexity $\mathcal{O}(n^3\log{}n)$ for a 2-D case. Over the years with efforts by Hupkens et al. \cite{hupkens2015faster}, Emmerich et al. \cite{emmerich2016multicriteria} and Yang et al. \cite{yang2017computing} the time complexity for 2-D and 3-D case has been reduced to $\mathcal{O}(n\log{}n)$. In this work an implementation of EHVI available on BoTorch based on estimating gradients using auto-differentiation is used as described by Daulton et al. \cite{daulton2020differentiable}. This exploits the high number of cores that are available with modern GPUs to make EHVI optimization fast and applicable to real-world scenarios. 

\subsection{Single-Objective, Multi-Fidelity Optimization}
\label{sec:2.3}

In many real-world optimization problems, multiple information sources with varying degrees of fidelity are available. These sources can provide different levels of accuracy, typically with a trade-off between data fidelity and cost. Integrating such multi-fidelity data into single-objective optimization algorithms is an essential way to improve the optimization process \cite{huang2006sequential,picheny2013quantile,swersky2013multi,klein2017fast,mcleod2017practical,zhang2017information}. 

Let us denote the input search parameters as $\bm{x}\in \mathcal{X}$ and the fidelity parameters as $\bm{s}\in \mathcal{S}$ where $\mathcal{X}$ and $\mathcal{S}$ are the input and fidelity spaces respectively. The goal is to build a surrogate model that incorporates information from $\bm{f(x,s)}$. The challenge in multi-fidelity optimization is to effectively balance the trade-off between information and cost while ultimately finding a global maximum at the target fidelity. This target fidelity usually corresponds to the highest fidelity which is also the most expensive information source.

One intuitive solution to this problem is the two-step approach proposed by Lam et al. \cite{lam2015}. In this approach, the selection of the next point to probe in $\bm{x}$ is done separately from the fidelity choice $\bm{s}$. To achieve this, Lam et al. use an Expected Improvement (EI) policy that is conditioned and evaluated at the target fidelity only. Lower fidelity measurements are implicitly incorporated into this process, as they affect the surrogate model at the highest fidelity. Once a suitable position has been identified in $\bm{x}$, the ideal fidelity for probing this point is chosen by comparing the predicted reduction in uncertainty or gain in knowledge with the computational cost involved.

An alternative approach is to combine both the selection of the next point and the weighting by the expected knowledge gain per unit cost in a single step. Notably, Max-Value Entropy Search (MES) and Knowledge Gradient (KG) acquisition functions can be adapted with minor changes for this kind of multi-fidelity optimization. For KG \cite{wu2018,wu2020}, the acquisition function is conditioned on the best value of the posterior mean at the target fidelity $s^*$ ($\max_{x\in\mathcal{X}}(\bm f(\bm x,s^*))$) rather than the best value of the posterior mean. In the case of MES, the mutual information between the maximum value $y^*$ at the highest fidelity and the data set is maximized. This gain of information is then divided by the computational cost that is a function of fidelity \cite{takeno2020}. The multi-fidelity MES acquisition function can be expressed as:

\begin{equation}
\begin{split}
\text{MF-MES}_n &= \frac{H(p_n(y^*|[x_{n},y_{n}],x_{n+1}))}{\text{cost}(s)} \\
&\quad- \frac{\mathbb{E}[H(p_n(y|[x_{n},y_{n}],x_{n+1},y^*))]}{\text{cost}(s)}.
\end{split}
\label{MF-MES}
\end{equation}

Similar multi-fidelity policies can be developed for other exploratory acquisition functions, such as the Upper Confidence Bound \cite{kandasamy2016,kandasamy2017}.

\section{Bayesian Multi-Objective and Multi-Fidelity Optimization}
\label{sec:3}

So far we have reviewed established techniques for multi-objective and multi-fidelity optimization. However, these methods are usually applied in isolation and, as we have seen, multi-objective optimization seeks to solve problems where the objective consists of several sub-objectives, whereas multi-fidelity optimization leverages data from multiple information sources with varying fidelity to optimize a single objective function. Both techniques offer unique benefits in their respective areas, but there are many real-world problems where the integration of both techniques would be beneficial. This leads us to the idea of joint multi-objective and multi-fidelity (MOMF) optimization, a method which incorporates both multi-objective and multi-fidelity aspects into a single optimization framework. Several compelling reasons motivate the pursuit of such an integrated optimization approach:

\begin{enumerate}
    \item \textbf{Efficiency:} Joint MOMF optimization efficiently leverages the strengths of both multi-objective and multi-fidelity optimization. The multi-objective aspect ensures diverse solution coverage, thoroughly exploring different areas of the solution space. Concurrently, multi-fidelity optimization minimizes the necessity of costly high-fidelity evaluations by using cheaper, lower-fidelity data sources where possible. This dual approach leads to a more effective balance between exploration and exploitation, thereby improving the overall optimization process.
    
    \item \textbf{Feasibility:} The complexity and cost of evaluating multiple objectives can often limit the feasibility of multi-objective optimization, particularly in real-world scenarios where such evaluations may involve physical experiments or high-fidelity simulations. By incorporating multi-fidelity optimization into the process, the MOMF approach leverages lower-fidelity, less costly models during the exploration phases. This considerably reduces the need for expensive high-fidelity evaluations, making the optimization process more manageable and feasible, even for scenarios where traditional multi-objective optimization would be prohibitively expensive or time-consuming.
    
    \item \textbf{Robustness:} The joint approach enhances the robustness of the optimization process by mitigating the risk of over-reliance on low-fidelity data. This risk is prevalent in single-objective multi-fidelity optimization and could introduce bias if the lower-fidelity models are inaccurate. By incorporating multiple objectives into the optimization process, the search strategy becomes more diversified, leading to a more robust and reliable optimization outcome.
    
    \item \textbf{Improved Decision Making:} MOMF optimization offers a more comprehensive understanding of the trade-offs involved in optimization tasks by allowing decision-makers to analyze the interplay between different objectives at varying levels of fidelity. This in-depth understanding facilitates more informed and holistic decision-making, ultimately leading to better optimization outcomes.
\end{enumerate}

Despite these apparent benefits, MOMF optimization remains an emerging area of research. A first paper reporting such an approach for discrete fidelity levels was recently published by Belakaria et al. \cite{belakaria2020}. Their MESMO (Max-value Entropy Search for Multi-Objective) Bayesian Optimization is based on the maximization of mutual information between the Pareto front and the search domain. Being mostly motivated by applications in neural network training, the approach from this paper is however limited to scenarios where higher fidelities yield higher objective values. The assumption that the objective values at lower fidelities are always upper-bounded by the values at the highest fidelity does not hold true for many use cases, such as numerical simulations of physical systems. Here we present two new approaches for Bayesian MOMF optimization that remedy these issues and are furthermore much simpler to implement for practitioners. We first introduce a holistic approach to combining the multi-objective and multi-fidelity optimization using trust. This method is able to select both the input parameters $x$ and the fidelity parameter $s$ jointly. To benchmark this approach, we also introduce a second method that selects the input parameters $x$ and the fidelity parameter $s$ sequentially. 

\begin{figure*}[t]
  \centering
  \includegraphics*[width=1\linewidth]{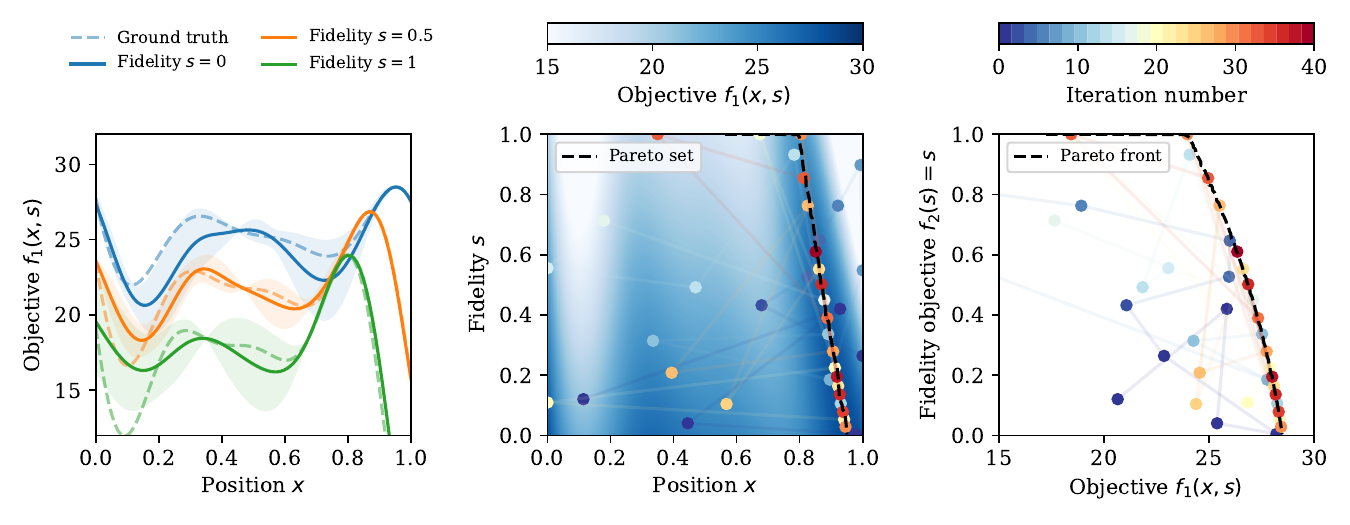}
  \caption{Multi-fidelity optimization via hypervolume improvement of a modified Forrester function. \textit{Left:} Mean and variance (shaded curve) of the fitted Gaussian process after 40 iterations of the optimizer. The true values are indicated as dashed lines. Note the small variance close to the maximum at all fidelity values. \textit{Center:} Map of the objective values $f_1(x,s)$ and sampled points colored according to the iteration number. This illustrates how the optimizer first explores at low fidelity and then moves along the Pareto set. \textit{Right: }Same optimization in the objective space, showing how the optimizer tries to increase the hypervolume, which is simply the area in 2D, below the Pareto front.}
  \label{MF-EHVI-1D}
\end{figure*}

\subsection{Trust-based Optimization}
\label{sec:3.2}

Our proposed optimization scheme hinges on the \emph{joint} optimization of a function and our \emph{trust} in the information source that yields the results. We use 'trust' to express our confidence level in the outcomes generated by the information source at various levels of fidelity. This formulation allows us to recast multi-fidelity optimization as a multi-objective problem, with trust $\theta(s)$ and output objectives $f(x,s)$ forming the two poles of optimization. Hence, our aim is to optimize the following function output:

\begin{equation}
    \bm f( x,s) = \begin{pmatrix}{ f({x},s)}
\\ \theta(s)
\end{pmatrix}.
\label{f_MOMF}
\end{equation}

So, what defines a trust objective $\theta(s)$? High fidelity sources, usually more costly, are intrinsically more trustworthy than their low fidelity counterparts due to their heightened accuracy and reliability. Therefore, trust grows monotonically with fidelity and could be defined as simply being equal to the fidelity itself, $\theta(s) = s$. A more rigorous approach may involve an actual measure of trust, e.g. linking the notion of trust to concepts such as mutual information. In these scenarios, the trust objective can be seen as an approximation that reflects the average mutual information shared across fidelities. For numerous circumstances, like simulations where the outputs at increasing fidelity converge, an appropriate trust curve follows approximately $\theta(s) \approx \tanh(s)$.

As discussed in Sec. \ref{sec:2.4}, we can optimize a problem of the form Eq.\ref{f_MOMF} using, for instance, Expected Hypervolume Improvement (EHVI). This acquisition function strives to increase the joint hypervolume encapsulated within the Pareto front of $f(x, s)$ and $\theta(s)$. 
Due due to the ascending nature of trust value, the optimizer tends to probe points with highest trust. But as in any multi-fidelity scenario where cost varies, we can introduce a cost-related penalizer (Sec. \ref{sec:2.3}). This ensures that the acquisition function invariably probes the point having the largest ratio between the expected hypervolume improvement and the associated cost. As a result, the acquisition function optimizes our knowledge of the Pareto front and Pareto set on a per-unit-cost basis. The entire structure of this approach for Trust-based Multi-Objective Multi-Fidelity (Trust-MOMF) is outlined in Algorithm \ref{Singlestep}.

Figure \ref{MF-EHVI-1D} exemplifies the optimization of a the 1D Forrester function by integrating information from lower fidelity data that incurs reduced cost. In this test case, trust is assumed linear ($\theta(s)=s$) and cost is modeled as $C(s)=\exp [a\cdot s]$, with $a=6$, rendering an evaluation at maximum fidelity ($s=1$) approximately 150 times costlier than at minimum fidelity ($s=0$). This cost function can be substituted with any other monotonically increasing function. A key assumption of Trust-MOMF is that the Pareto set, which underlies the Pareto front, belongs to a similar region in the search space, thereby facilitating efficient translation of information across fidelities. It's important to note that this method, akin to all multi-fidelity approaches that leverage knowledge transfer, becomes less efficient if the Pareto set experiences significant shifts between fidelities.

Finally, we would like to point out that this method can be expanded to include multiple fidelity dimensions $\bm s^m$ where $m$ represents the number of fidelity dimensions. This is especially useful for multi-dimensional numerical models with individual resolution parameters. Depending on the problem, these individual fidelity dimensions can either be managed via a single, unified trust objective or treated as separate dimensions within the optimization problem. However, it should be noted that the EHVI algorithm does not scale efficiently to many dimensions, and the sequential MOMF introduced in the next section may be a more suitable alternative when numerous fidelity dimensions need to be optimized individually.

\begin{algorithm}[t]
\caption{Trust-based Multi-Objective Multi-Fidelity Optimization (Trust-MOMF)}
\label{Singlestep}
\SetKwInOut{Input}{Inputs}
\Input{Probed Dataset $D=(\bm x_{n-1},\bm y_{n-1})$, Models $GP_1,GP_2,...,GP_k$, each function has a continuous fidelity $s$, $C_{total}$ represents the total available cost}
Generate Initial data $D=(\bm x_{n-1},\bm y_{n-1})$ and build the surrogate models $GP_1,GP_2,...,GP_k$ for objective functions $f_1,f_2,...,f_{k}(\bm x)$. Note that the last function $f_k$ is a fidelity-related objective\;
Generate MC-samples for the estimation of the Expected Hypervolume Improvement acquisition function\;
\While{$C_i<C_{total}$}{
    $x_i,s_i \leftarrow \text{argmax}_{x\in \mathcal{X}}[\text{MF-EHVI}(\bm x|D,s=1)]$ \Comment{Hypervolume/cost optimization}\;
    $y_i \leftarrow \text{Problem}(x_i,s_i)$ \Comment{Evaluating objective function at $x_i$ and fidelity $s_i$}\;
    $D_{n} \leftarrow \{x_i,y_i\}$ \Comment{Updating dataset with the new input-output pair}\;
    Update models $GP_1,GP_2,...,GP_k$\;
}
\end{algorithm}

\subsection{Sequential Optimization}
\label{sec:3.1}

While we have so far discussed the benefits and implementation of joint Trust-based Multi-Objective Multi-Fidelity (Trust-MOMF) optimization, an alternative approach worth exploring is a sequential version of the MOMF optimization. This sequential optimization scheme, inspired by the work of Lam et al. \cite{lam2015} on single-objective optimization, separates the selection of the next position to probe and the fidelity choice into two distinct steps, thus bringing a different perspective to multi-fidelity optimization. In the following sections, we will present and discuss this method in depth, outlining its main principles, operational procedures, and potential benefits. This sequential MOMF approach will also serve as a benchmark to evaluate the extent to which our problems benefit from a joint optimization strategy.

\begin{algorithm}[t]
\caption{Sequential Multi-Objective Multi-Fidelity Optimization (Seq. MOMF)}
\label{Two-Step}
\SetKwInOut{Input}{Inputs}
\Input{Probed Dataset $D=(\bm x_{n-1},\bm y_{n-1})$, Models $GP_1,GP_2,...,GP_k$, each function has a continuous fidelity $s$, $C_{total}$ represents the total available cost}
Generate Initial data $D=(\bm x_{n-1},\bm y_{n-1})$ and build the surrogate models $GP_1,GP_2,...,GP_k$ for objective functions $f_1,f_2,...,f_{k}(\bm x)$. Also build another surrogate model for a scalarized objective to be given to Fidelity selector\;
Generate MC-samples for the estimation of the Expected Hypervolume Improvement acquisition function\;
Generate a candidate set of points that discretizes the input space where MF-MES will be calculated\;
\While{$C_i<C_{total}$}{
    $x_i \leftarrow \text{argmax}_{x\in \mathcal{X}}[\text{EHVI}(\bm x|D,s=1)]$ \Comment{Highest fidelity hypervolume optimization}\;
    $s_i \leftarrow \text{argmax}_{s\in \mathcal{S}}[\text{MF-MES}(x|D,x_i)]$ \Comment{Selecting Fidelity using MF-MES}\;
    $y_i \leftarrow \text{Problem}(x_i,s_i)$ \Comment{Evaluating objective functions at $x_i$ and fidelity $s_i$}\;
    $D_{n} \leftarrow \{x_i,y_i\}$ \Comment{Updating dataset with the new input-output pair}\;
    Update models $GP_1,GP_2,...,GP_k$ and the Fidelity selector GP\;
}
\end{algorithm}

Here we are going to use EHVI to suggest the next point in the input space $x_{n+1}$ and combine it with the MF-MES method that maximizes information gain on a scalarized objective given the fidelity point $s_{n+1}$. The scalarization is done by summing all the objectives with equal weights to avoid any preference between objectives. This assumes that all the objective functions change at a similar scale which was insured in this study by scaling all the test functions in the range [0,1]. The method is summarized in the Algorithm \ref{Two-Step}. We call this method a Sequential multi-objective multi-fidelity (Seq. MOMF) optimization because the multi-objective optimization of $\bm x$  and the selection of the fidelity point $\bm s$ is performed sequentially.  It should be noted that one can use any multi-fidelity acquisition function (such as KG or UCB, see Section \ref{sec:2.3}) for selecting the fidelity parameter $\bm s$ in this procedure. 

As we will discuss in Section \ref{Benchmark}, this rather straightforward implementation of a MOMF policy already provides a significant speedup compared to pure multi-objective optimization. An advantage of this scheme is the negligible computational overhead compared to pure MO optimization, as the information gain across fidelities only needs to be calculated at the already selected candidate point.

\subsection{Comparison and Benchmark}\label{Benchmark}
\label{sec:3.3}
In this section we will describe the results of our proposed Trust-MOMF and sequential MOMF algorithms on synthetic test functions. All benchmarking is performed by modifying existing implementations of MES and EHVI in the BoTorch package \cite{balandat2020} to the multi-objective, multi-fidelity problem. \\

\textit{Test functions.} To assess the performance of the methods and estimate the cost reduction factors, we use multi-fidelity modifications of the popular maximizing Branin-Currin (2-D) and Park (4-D) test functions. 
The function definitions for the Branin-Currin and Park are given in the Appendix. The cost function for the different fidelities is modeled as an exponential function of the form $C(s)=\exp [a\cdot s]$ with $a=5.7$ to result in a ratio of about 120:1 between the highest ($s=1$) and lowest ($s=0$) fidelity. The number of iterations for the MOMF algorithms was fixed to 120 while the multi-objective single-fidelity optimization referred to as MO ran for 80 iterations. For the MO optimization the total cost was 9600 while the MOMF algorithms stopped at variable costs ranging from 1500 to 4000. \\

\textit{Initialization.} Both MOMF algorithms are initialized with five starting points that are randomly distributed in the input search space. The single-fidelity MO optimizer is initialized with a single point at the highest fidelity $s=1$ and thus starts at an initial cost of $C(1)\simeq 120$. Meanwhile, the fidelity of initial points for the MOMF optimizers is drawn from a cost-aware probability distribution of the form $p(s)\propto 1/C(x)$, resulting on average in a five-times lower initialization cost. As the choice of initial points can influence the performance of the optimization, we run each optimizer 10 times with different initialization to attain more robust statistics on the convergence of the algorithms. Note that in the first two iterations due to lack of training points for the GP model we obtain a large set of non-dominated points making this calculation computationally expensive, thus without any loss of information this calculation is performed after four iterations. This is the reason for the graph starting from a cost of 480 for MO optimizations in the following figures.\\ 

\textit{Hypervolume calculation.} For hypervolume calculation, the points taken during the optimization run are used as training inputs for a GP model while a random input sample of 10000 points at the highest fidelity are taken as test inputs. From these 10000 points a set of non-dominated points and consequently the hypervolume is calculated at each iteration step. Please note that the estimated hypervolume may occasionally decrease between iterations (see for instance sequential MOMF at cost $\sim 600$ in Fig.4). This is because we calculate the hypervolume via the GP and not only via the dominating measurements at maximum fidelity. When a point is added to the training inputs, this can have the effect of decreasing the hypervolume temporarily since the model might predict differently at other places of the 2-D space. After subsequent learning of the objective function this effect is minimized since the GP model becomes more certain of its predictions. \\

\begin{figure*}[t]
    \centering\
    \includegraphics[width=0.97\textwidth]{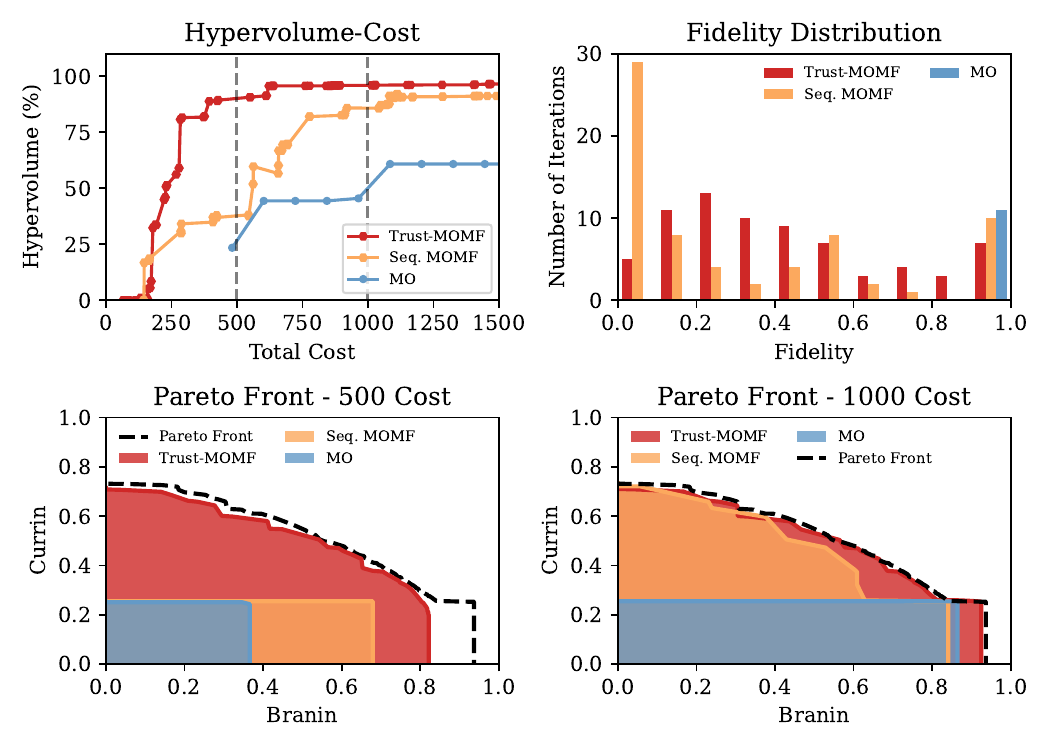}
    \caption{Benchmark with 2-D-Branin-Currin problem. \textit{Top Left: }Hypervolume of a single representative trial expressed as a percentage of the total hypervolume versus total cost for both MOMF versions and single-fidelity MO. \textit{Top Right: }The number of points taken at different fidelites for the representative trials shown on the top left. The Trust-MOMF takes more iterations at intermediate fidelities when compared with sequential MOMF which has relatively high peaks at fidelity 0 and 1. \textit{Bottom Left: }The Pareto front for each of the three algorithms for the same trial at a cost of 500 (indicated by the dashed line in the figure on the top left). The dashed black line represents an estimated Pareto front calculated from 10000 random points. The Trust-MOMF has already found values in the trade-off region and close to the individual maxima of the Branin and Currin functions. The sequential MOMF is approaching the maximum of the Branin function but both sequential MOMF and MO algorithms have yet to explore the trade-off points. \textit{Bottom Right: }The Pareto front for a cost of 1000 cost for a single representative trial, clearly showing how the Trust-MOMF has reached an accurate representation of the estimated Pareto front while the conventional, single-fidelity MO algorithm has only discovered a small region of the Pareto front at this cost.}
    \label{HV_costBC}
\end{figure*}

\subsubsection{Results}\label{Results}

\textit{Branin-Currin test function.} In Figure \ref{HV_costBC} an optimization of the multi-fidelity versions of Branin and Currin \cite{dixon1978,currin1991} functions is shown. A single representative trial that was close to the mean hypervolume as a function of cost is depicted on the top left. The regular steps along the cost axis for MO optimization indicate a fixed cost at the highest fidelity, while for both the MOMF optimizations it can be seen that step sizes are irregular. The MOMF algorithms take a few points at intermediate fidelities before taking a high fidelity point. This can also be seen in top right of the figure where the distribution of selected fidelities is shown for each algorithm. Interestingly, we observe a different behavior between the two MOMF algorithms. The sequential MOMF takes considerably more points at the lowest fidelity, while the Trust-MOMF takes much more intermediate fidelity points, possibly because of the joint optimization of both input and fidelity space. 

\begin{figure*}[t]
    \centering
    \includegraphics[width=0.98\textwidth]{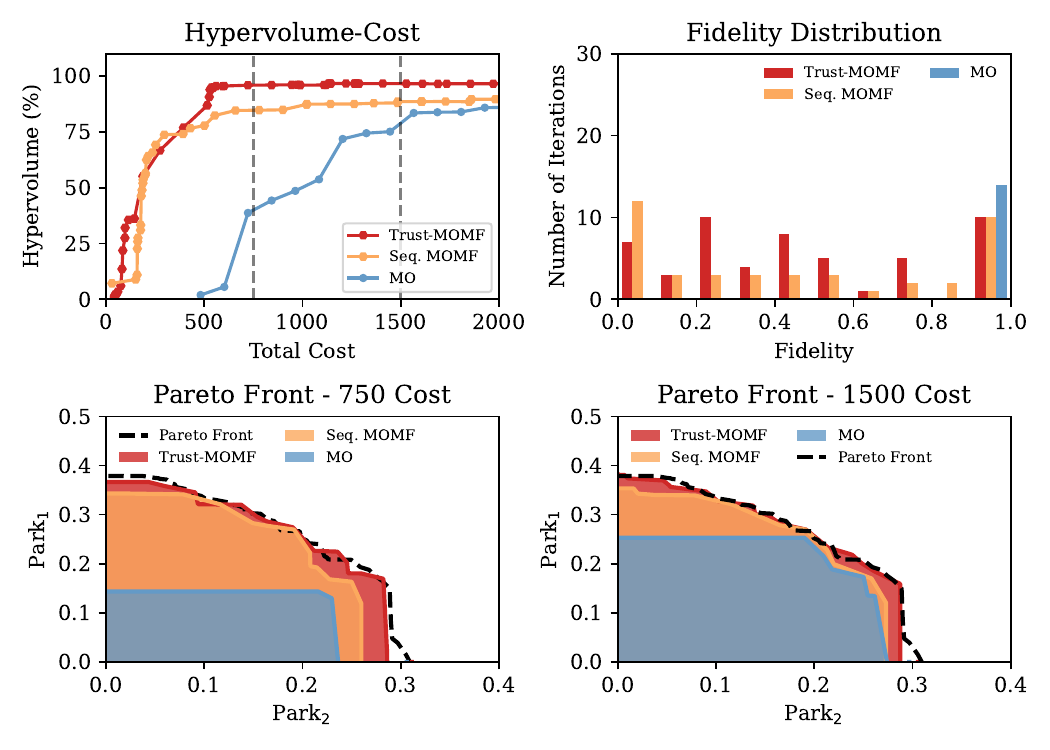}
    \caption{Benchmark with 4-D-Park$\bm{_{1,2}}$ problem. \textit{Top Left: }Hypervolume of a single representative trial expressed as a percentage of the total hypervolume versus total cost for both MOMF versions and single-fidelity MO. \textit{Top Right: }The number of points taken at different fidelites for the representative trials shown on the left. The Trust-MOMF in this case has a higher number of points taken at the highest fidelity. This is because once it has converged it takes 6 points at the highest fidelity to increase hypervolume. The sequential MOMF as seen in Branin-Currin case takes less intermediate fidelity points when compared to the Trust-MOMF. \textit{Bottom Left: }The Pareto front for each of the three algorithms for the same trial at a cost of 750 (indicated by dashed line in the figure on the top left). The area represents the amount of Pareto front covered by each algorithm. The Trust-MOMF has already converged to almost $95\%$ of the total hypervolume. The sequential MOMF also converged but to a lower overall hypervolume. The MO optimization at this cost has only found maximum of Park 2 function.\textit{Bottom Right: }The Pareto front for a cost of 1500 cost showing little changes in both MOMF Pareto fronts, but a better coverage for the MO Pareto front. At this cost the MO optimization still has not reached the hypervolume that the Trust-MOMF reached at a cost of 750.}
    \label{HV_costPP}
\end{figure*}

The bottom part of Figure \ref{HV_costBC} depicts the behaviour of the Pareto front at two different costs. The black-dashed line represents the true Pareto front calculated using 50000 random points. Here it can be seen that the Trust-MOMF already has a good coverage of the trade-off region and the maximum of the Currin function. The MO and sequential MOMF algorithms at a cost of 500 have much less hypervolume coverage. At a cost of 1000 the MO optimization has optimized the Branin function but still has not explored the trade-off region. The sequential MOMF at a cost of 1000 is in a similar state as that of Trust-MOMF at a cost of 500. Meanwhile, the Trust-MOMF has reached near $97\%$ hypervolume coverage.

\begin{figure*}[t]
    \centering
    \includegraphics[width=1\textwidth]{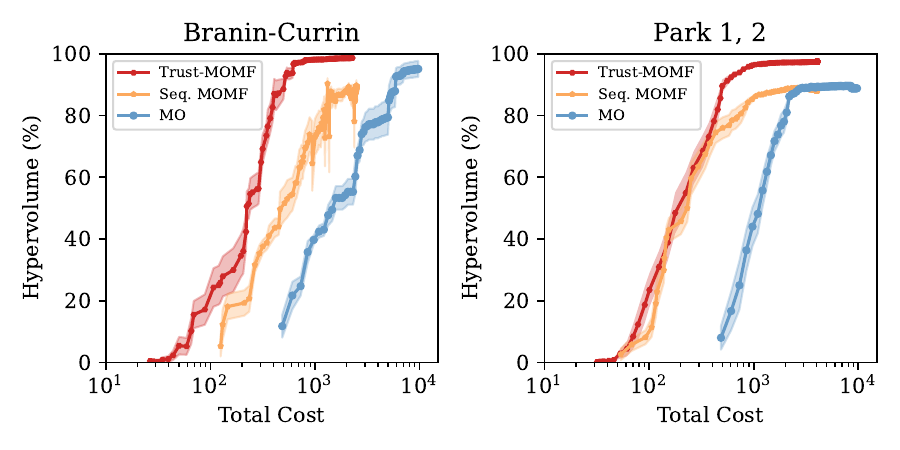}
    \caption{Mean Hypervolume with the shaded region indicating the standard deviation of 10 trials for Branin-Currin and Park functions. \textit{Left: }Mean Hypervolume as a percentage of the total for 10 trials of Branin-Currin optimization is illustrated. The total cost is in the log scale to depict the large differences in the cost. The dips seen in the curve around 900 are due to the method of calculating the hypervolume described earlier. When considering convergence to $90\%$ hypervolume an order of magnitude cost difference is seen between Trust-MOMF and MO optimizations. \textit{Right: }The mean hypervolume of 10 trials for the Park functions is shown. Here again considering a convergence to $90\%$ almost an order of magnitude advantage is seen between the Trust-MOMF and MO optimizations.}
    \label{MeanHV}
\end{figure*}

For the estimation of the cost advantage we use the average hypervolume cost curves from ten runs (shown at the left in Figure \ref{MeanHV}). Taking $90\%$ convergence as a threshold, there is an order of magnitude cost advantage for the Trust-MOMF over the MO optimization. The MO algorithm converged to $90\%$ at about a cost of 6000 while the Trust-MOMF reached the same hypervolume at a cost of 530, resulting in a cost reduction factor of $\sim 11$. Moreover the Trust-MOMF converged to a higher hypervolume percentage ($99\%$) when compared with final convergence of $94\%$ for MO optimization. \\

\textit{Park test function.} Figure \ref{HV_costPP} depicts a representative trial run results of the same three algorithms for the optimization run of modified Park functions \cite{park1991tuning}. The top left sub-figure shows the hypervolume percentage covered versus total cost for the MOMF and MO optimization runs. The Park problem shows a similar behavior regarding the fidelity distribution, i.e., ~the sequential MOMF concentrates on points at minimum ($s=0$) and maximum  ($s=1$) fidelity, while the Trust-MOMF takes more intermediate fidelity points. On the bottom two plots of Figure \ref{HV_costPP} we see the evolution of the Pareto fronts for the 3 optimizers. The Trust-MOMF already at a cost of 500 has converged to a hypervolume coverage of $94\%$, hence its Pareto front at a cost of 750 is close to the true Pareto front. The sequential MOMF also has started exploring the trade-off region and consequently pushes out the Pareto front slightly up to a cost of 1500, as shown on the right. The MO optimization at the cost of 750 has found points close to Park 2 maximum and thus explores the trade-off region at a cost of $1500$. The cost advantage estimation is again done using the mean hypervolume versus total cost curves generated using $10$ trials as shown in Figure \ref{MeanHV}. The MO algorithm converged to a hypervolume of $89.8\%$ at a cost of about $7600$, whereas the Trust-MOMF reached $90\%$ hypervolume coverage at a cost of about 560. This results in a cost reduction factor of about $13$, similar to the Branin-Currin problem. 

\subsubsection{Discussion}

For both test functions considered we have observed significant cost reduction in finding the Pareto front of each problem using either the joint trust-based or sequential MOMF optimization. The joint optimization generally outperforms sequential optimization, hinting at its better use of available information within the joint search domain. It should be noted that the cost reduction is intrinsically linked to the cost ratio between lowest and highest fidelity. In our examples, this ratio was 1:120 and thus, the highest possible cost reduction by taking only lowest fidelity data points would be 120. Averaging the fidelity of the Trust-MOMF over $10$ trials and $150$ iterations for both Branin-Currin and Park functions yields $\overline{s}\simeq 0.30$ and $\overline{s}\simeq 0.32$, respectively. This results in an average cost of $4.3$ and $4.6$ per iteration, while the conventional MO optimization is fixed to a cost of $120$. Thus, a cost reduction of 26 is the maximum that could be achieved if the information gain were the same at all fidelity levels. However, this is generally not the case and low-cost approximations do normally not carry as much information as the highest fidelity. In our case, we see a cost reduction of about half this best possible value, i.e., a cost reduction factor of $13$ in both considered test cases. The cost reduction factor can be even higher when the ratio between the cost at highest and the lowest fidelity is increased. For instance, at a maximum cost ratio of 1200:1 for the Branin-Currin problem we measure a cost reduction factor of 44 for convergence to 90\% of the hypervolume. 

\begin{figure*}[t]
    \centering
    \includegraphics[width=0.98\textwidth]{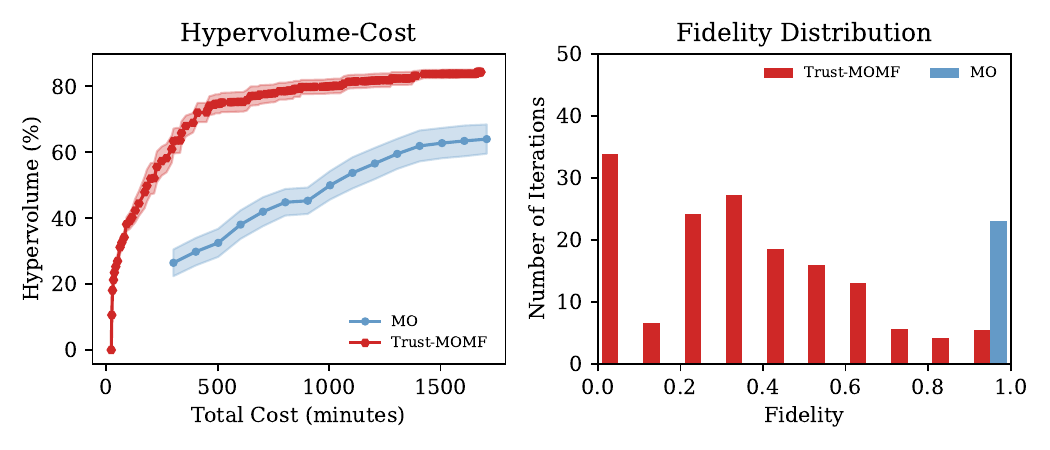}
    \caption{Benchmark with PIC simulations of laser wakefield accelerator using a 4-D input space. \textit{Left: } Mean hypervolume from 10 trials for the Trust-MOMF and the MO optimization expressed as a percentage vs the time taken in minutes. The shaded region indicates the standard deviation calculated from the 10 trials. For each run the computational budget was 30 hours. The sequential MOMF algorithm was excluded because of limited computational and time resources when optimizing with simulations.  \textit{Right: } The number of points taken at different fidelities for both the Trust-MOMF and the MO optimization for a single trial.}
    \label{HV_costFBPIC}
\end{figure*}

\subsection{Application to Particle-In-Cell Simulations of Laser-Plasma Acceleration}

In this section we present an application  from computational physics that makes full use of the new joint MOMF acquisition function to speed up optimization.

One promising application area for MOMF optimization is computational physics. Here we are going to discuss an example from the realm of laser-plasma interactions, which are governed by the Vlasov-Maxwell equations. This kind of system is often simulated using the particle-in-cell (PIC) method \cite{Vay.2016}, where the Maxwell equations are solved on a numerical grid, and the Vlasov equation is solved continuously using a set of macroparticles. The need to compute the trajectories of millions of simulated particles and their fields makes PIC simulations computationally quite costly. For our example, we used the FBPIC which is an open-source python code frequently used for simulations of laser or plasma wakefield accelerators \cite{Lehe.2016}. FBPIC is a GPU-based code that uses several approximations, making it a relatively ``lightweight'' PIC code, with typical runtimes on the order tens of minutes for test runs and a several hours or days for production runs on a single high-end GPU. Extensive scans at production run resolution are hence prohibitively expensive, making this an ideal test case for the use of a multi-fidelity optimization. 

In our example the laser wakefield accelerator generates quasi-mono energetic electron beams with GeV-level energies and charge of up to a few hundred picocoulomb (pC), similar to the conditions reported in Götzfried et al. \cite{Gotzfried.2020}. The optimization goal was to produce beams near a target energy of 300 MeV with low bandwidth and high charge plasma electron density, as well as the shape of the plasma profile (upramp and downramp length). The output objectives are the total charge in the beam, the distance of mean energy to the target energy and the standard deviation of the energy spectrum. The distance to the target energy and the standard deviation are minimization objectives while the charge is to be maximized. This makes the optimization a $4 \times 3$ D optimization, making it difficult to depict the Pareto surface. Since these are expensive simulations we also only use the Trust-MOMF to optimize and compare it with MO optimization. The cost function for the FBPIC simulations considered here scales as an exponential of the form $C(s)=\exp [a\cdot s]$ with $a=6$ resulting in a ratio of about 150:1. The computational budget for each run was capped to 30 hours. In Figure \ref{HV_costFBPIC} we see that the Trust-MOMF algorithm outperforms the MO optimization by converging to a hypervolume of about $85\%$ while the MO optimization in this budget converged to a hypervolume of about $63\%$. The MO optimization was run for 8 more hours but still it failed to reach a hypervolume of more than $73\%$. Also shown in Figure \ref{HV_costFBPIC} is the fidelity distribution which shows that the Trust-MOMF takes a large amount of simulations with a fidelity of less than $0.5$, thus efficiently making use of these faster, low-resolution simulations. Comparing the costs we see that the Trust-MOMF in this case provides a cost reduction factor of about $6$, which similar to the gain we previously estimated with test functions. These encouraging results confirm the usefulness of joint MOMF for expensive numerical simulations in physics.

\section{Summary and Outlook}
\label{sec:4}

In conclusion, we have successfully demonstrated the practical implementation of simultaneous multi-fidelity and multi-objective optimization, achieving substantial cost reductions - over an order of magnitude - for test problems with a cost ratio of 1:120. Our Trust-MOMF method showed superior convergence towards a higher hypervolume compared to single-fidelity multi-objective optimization approaches. We furthermore observed that the joint Trust-MOMF policy consistently outperforms the sequential MOMF strategy, indicating the beneficial use of joint information across objectives and fidelities. We also discussed the theoretical upper limit for the cost reduction factor, which can be amplified further by widening the cost ratio between the highest and lowest fidelity levels.

The significance of these findings lies in the substantial cost-efficiency MOMF offers, particularly for applications aiming to optimize multiple objectives where access to lower fidelity data is available. Our approach, which builds upon existing acquisition functions, is easily implementable within optimized Bayesian optimization frameworks such as BoTorch \cite{balandat2020}. It can be seamlessly extended, for instance, to batch optimization.

Our work contributes a key optimization tool for complex problems in fields such as physics and engineering, where simulations with varying degrees of accuracy are commonly used. Moreover, it holds potential for any domain where there is an advantage in optimizing different objectives while also having access to less costly evaluations. The flexibility of our approach is underscored by its adaptability to any other multi-objective, single-fidelity acquisition function - a trust objective can be introduced and the acquisition value penalized by cost. Additionally, our MOMF methods can be extended to include multiple fidelity dimensions, offering the prospect of efficient optimization over different resolution parameters in numerical simulations.

Code samples for the benchmark cases are available online for further exploration and experimentation. As we look to the future, we anticipate the increasingly broad application and continued evolution of multi-fidelity and multi-objective optimization techniques.\\

\textbf{Acknowledgements.} This work was supported by the DFG through the Cluster of Excellence Munich-Centre for Advanced Photonics (MAP EXC 158), TR-18 funding schemes and the Max Planck Society. It was also supported by the Independent Junior Research Group ``Characterization and control of high-intensity laser pulses for particle acceleration", DFG Project No.~453619281. F.I. is part of the Max Planck School of Photonics supported by BMBF, Max Planck Society, and Fraunhofer Society.\\

{\textbf{Data availability. } The code is publicly available in the linked repository \href{https://github.com/PULSE-ML/MOMFBO-Algorithm}{repository} and relevant functions have been integrated into the BoTorch repository. Further material such as simulation data are provided upon reasonable request.}

\newpage
\appendix
\onecolumngrid
\section*{Appendix A}
\section*{Notation}

\begin{center}
\begin{tabular}{c|c}
 \hline \\
 \textbf{Notation} & \textbf{Description} \\[3mm]
 \hline
 \textbf{$\bm x$, $\bm y$ , $\bm f$, $\bm s$} & bold notation for vectors  \\[0.5mm]
 \hline
 $f_1,f_2,...,f_k$ & k number of objective functions \\[0.5mm]
 \hline
 \textbf{$\bm x$, $\bm y$} & input and output vector  \\[0.5mm]
 \hline
 \textbf{$\bm f$, $\bm s$} & vector of objective functions and fidelity vector \\ [0.5mm]
 \hline
 $y_k^s$& value of kth function evaluated at fidelity $s$\\[0.5mm]
 \hline
  $\lambda_k^s$ & cost of evaluating kth function at fidelity s \\[0.5mm]
  \hline
  $\mathcal{Y^*}$ & true Pareto front at the highest fidelity\\[0.5mm]
\end{tabular}
\end{center}

\section*{Definitions}

\begin{definition}
A point $p_1=(y_1,y_2)$ in the 2D output space is defined as \emph{non-dominated} if there does not exist another point $p_2^{'}=(y_1^{'},y_2^{'})$ such that $p_2^{'}$ has equal or superior values for all objectives and a strictly superior value for at least one objective. Formally, there is no $p_2^{'}$ that satisfies ($y_1^{'}\geq y_1 \land y_2^{'}\geq y_2$) with at least one strict inequality.
\label{domination}
\end{definition}

\section*{GP Kernel choice}

Many studies motivated by applications such as neural network training use an exponential decay kernel and a zero mean prior by default. In this paper, we have $k$ number of $GP_1,GP_2,...,GP_k$ models modelling $k$ objective functions $f_1,f_2,...,f_{k}(\bm x)$. Each of these functions also has an input fidelity parameter $s$ that implies greater accuracy with higher values. We have chosen a Matern 5/2 kernel to model the fidelity dependence, which has shown significantly better performance on our test and application cases than the exponential decay kernel.

\section*{Test Functions}

In this section we describe the analytical functions used to benchmark the performance of our algorithm. As this is one of the first studies on combined multi-objective and multi-fidelty optimization, we could not directly use test functions from the literature but had to modify them to incorporate an additional fidelity input dimension and exhibit trade-off behavior between the objectives.

\subsubsection*{\bf Modified Multi-Fidelity Forrester Function}

The original Forrester function is
$$f(x) = (6 x - 2)^2 \cdot \sin(12 x - 4) + 7.025$$

and we use the following modified version as multi-fidelity reference case:
$$f(x,s) = D(s)\cdot\left[E-g(x,s)\right]$$
where
$$g(x,s)=A(s)\cdot f[x-0.2(1-x\cdot s)] + B(s)\cdot(x-0.5) - C(s)$$
with $A = 0.5+0.5s$, $B = 2 - 2s$, $C =  5s-5$, $D = 1.5 - 0.5s$ and $E=25$. The most important differences to other multi-fidelity versions of this function are that it contains a fidelity- and position-dependent shift $\tilde x(x,s) =x-0.2(1-x\cdot s)$, is inverted for maximization ($E-g(x,s)$ term), the value of the maxima along the fidelity is decreasing (D(s) term) and the maxima are continuously connected from at low to high fidelity.

\subsubsection*{\bf Modified Multi-Fidelity, Multi-objective Branin-Currin Function}
Two popular functions used for optimization benchmarking are Branin-Currin functions which were also modified. The usual form of the Branin function is 
$$B(\bm x)=a(x_2-bx_{1}^2+cx_1-r)^2+p(1-t)\cos{(x_1)}+p,$$ 
where values of the constants were taken to be $a=1$, $b=5.1/(4\pi^2)$, $c=5/\pi$, $r=6$, $p=10$, $t=1/(8\pi)$ and the form of Currin function is 
$$C(\bm x)=\left[1-\exp{\left(-\frac{1}{2x_2}\right)}\right]\frac{2300x_{1}^3+1900x_{1}^2+2092x_{1}+60}{100x_{1}^3+500x_{1}^2+4x_1+20}.$$
Both of these functions were modified to make the range of input and output values to be between 0 and 1, i.e., $x_i,y_i\in [0,1] \: \forall i=[1,2]$. The modified form that was used for the Branin is 
$$B(\bm x,s)=-[a(x_{22}-b(s)x_{11}^2+c(s)x_{11}-r)^2+p(1-t(s))\cos{(x_{11})}+p]$$ 
where $x_{11}=15(x_1)-5$, $x_{22}=15x_2$, $a=1$, $b(s)=5.1/(4\pi^2)-0.01(1-s)$, $c(s)=5/\pi-0.1(1-s)$, $r=6$, $p=10$, $t(s)=(1/(8\pi))+0.05(1-s)$ was used. The $x_{11}$ and $x_{22}$ are used to scale the original domain of the Branin to [0,1]. The main difference is the addition of fidelity parameter s which for a value of 1 yields the original Branin function and since we are maximizing the problem a minus sign is added. The modified form for the Currin function is 
$$C(\bm x,s)=-\left[\left[1-(0.1)(1-s)\exp{\left(-\frac{1}{2x_2}\right)}\right]\frac{2300x_{1}^3+1900x_{1}^2+2092x_{1}+60}{100x_{1}^3+500x_{1}^2+4x_1+20}\right]$$
where again the main difference is the addition of fidelity term $1-s$ and the addition of a minus sign for maximization.

\subsubsection*{\bf Modified Multi-Fidelity Multi-Objective Park Functions}
For a benchmark of the multi-objective multi-fidelity problem in higher dimensions, multi-fidelity versions of Park 1 and Park 2 functions were used. The original form of the Park functions is 
$$P_1(\bm x) = \frac{x_1}{2}\left[\sqrt{1+(x_2+x_{3}^{2})\frac{x_4}{x_{1}^{2}}}-1\right]+(x_{1}+3x_4)\exp{[1+\sin({x_3})]}$$
$$P_2(x)=\frac{2}{3}\exp{(x_1+x_2)}-x_4\sin{(x_3)}+x_3$$

We modified the above two Park functions adding a fidelity dimension ($s$). To achieve a reasonable Pareto front for optimization, the two functions were also slightly modified. The location of the Pareto set was also modified to not have all the optimizing points in the corners of the 4-D hypercube. A last modification is shifting the Pareto front of the Park functions by subtraction to place a higher importance on the trade-off region. The final form of the two modified Park functions was 
$$P_1(\bm x^{'},s) = A(s) \left[T_1+T_2-B(s)\right]/22-0.8$$
$$T_1=\left[\frac{x_1+0.001(1-s)}{2}\right].\left[\sqrt{1+(x_2+x_{3}^{2})\frac{x_4}{x_{1}^{2}}}\right]$$
$$T_2=(x_{1}+3x_4)\exp[{1+\sin({x_3})}]$$
$$P_2(\bm x^{'},s)=A(s)\left[5-\frac{2}{3}\exp{(x_1+x_2)}-(x_4)\sin{(x_3)A(s)}+x_3-B(s)\right]/4-0.7$$
where $A(s)=(0.9+0.1s)$ and $B(s)=0.1*(1-s)$. Both Park functions now contain a fidelity parameter $s$. These Park functions are evaluated on a transformed input space $$[x_1,x_2,x_3,x_4]\rightarrow [1-2(x_1-0.6)^{2},x_2,1-3(x_3-0.5)^{2},1-(x_4-0.8)^{2}].$$

\twocolumngrid

\end{document}